\documentclass[letterpaper, 10 pt, conference]{ieeeconf} 
\IEEEoverridecommandlockouts

\pdfoutput=1
\usepackage{url}

\usepackage{graphicx}

\usepackage{amsmath,amssymb,amsfonts}

\usepackage{hyperref}

\usepackage{bm}
\usepackage{xcolor}
\usepackage{multirow}
\usepackage{paralist}
\usepackage{wrapfig}
\usepackage[capitalise]{cleveref}
\usepackage{siunitx}
\usepackage{booktabs}
\usepackage{caption}
\usepackage{pgfplots}
\pgfplotsset{compat=newest}
\usepackage{graphicx}
\usepackage{etoolbox,siunitx,booktabs,threeparttable,multirow,array,graphicx}
\usepackage[normalem]{ulem}
\usepackage{makecell}
\usepackage{siunitx}
\robustify\bfseries
\robustify\uline
\robustify\tnote

\usepackage[utf8]{inputenc}
\usepackage{pgfplots}
\DeclareUnicodeCharacter{2212}{−}
\usepgfplotslibrary{groupplots,dateplot}
\usetikzlibrary{patterns,shapes.arrows}
\pgfplotsset{compat=newest}

\usepackage[style=ieee,citestyle=numeric-comp]{biblatex}
\usepackage[ruled,linesnumbered]{algorithm2e}

\usepackage[nolist,nohyperlinks]{acronym}
\begin{acronym}
\acro{MAV}{Micro Aerial Vehicle}
\newacroindefinite{MAV}{an}{a}
\acro{OMAV}{Omnidirectional Micro Aerial Vehicle}
\newacroindefinite{OMAV}{an}{an}
\acro{UAV}{Uncrewed Aerial Vehicle}
\acro{ESDF}{Euclidian Signed Distance Field}
\acro{IMU}{Inertial Measurement Unit}
\acro{RMP}{Riemannian Motion Policy}
\acro{EDF}{Euclidean Distance Field}
\newacroindefinite{IMU}{an}{an}
\acro{VTOL}{vertical take-off and landing}
\acro{SLAM}{Simulateous Localization and Mapping}
\acro{MDE}{Monocular Depth Estimation}
\acro{SfM}{Structure from Motion}
\acro{SOTA}{state-of-the-art}
\acro{LiDAR}{Light Detection And Ranging}
\acro{TOF}{Time Of Flight}
\acro{FMCW}{frequency-modulated continuous-wave}
\acro{CNN}{Convolutional Neural Network}
\acro{CFAR}{constant false alarm rate}
\acro{SNR}{signal-to-noise ratio}
\acro{GNSS}{Global Navigation Satellite System}
\end{acronym}

\begin{document}

\newcommand{\todo}[1]{}
\renewcommand{\todo}[1]{{\color{red} TODO: {#1}}}

\title{Radar Meets Vision: Robustifying Monocular Metric Depth Prediction \\for Mobile Robotics}
\author{Marco~Job, 
        Thomas~Stastny,
        Tim~Kazik,
        Roland~Siegwart, 
        Michael~Pantic
\thanks{All authors are with the Autonomous Systems Lab, ETH Z\"urich, Switzerland %
        {\tt\footnotesize \{mjob,tstastny,tkazik,rsiegwart, mpantic\}@ethz.ch}.}
        \thanks{This work has been supported by a Swiss Polar Institute Technogrant, the Armasuisse Research Grant No 4780002580, and by a ETH RobotX research grant funded through the ETH Zurich Foundation. We also thank Wingtra AG for the permission to use the Rural Fields, Mountains (Windpark), and Road Corridor datasets.}
}

\maketitle%
\thispagestyle{empty}
\pagestyle{empty}

\begin{abstract}
Mobile robots require accurate and robust depth measurements to understand and interact with the environment.
While existing sensing modalities address this problem to some extent, recent research on monocular depth estimation has leveraged the information richness, yet low cost and simplicity of monocular cameras.
These works have shown significant generalization capabilities, mainly in automotive and indoor settings.
However, robots often operate in environments with limited scale cues, self-similar appearances, and low texture.
In this work, we encode measurements from a low-cost mmWave radar into the input space of a state-of-the-art monocular depth estimation model.
Despite the radar's extreme point cloud sparsity, our method demonstrates generalization \emph{and robustness} across industrial and outdoor experiments.
Our approach reduces the absolute relative error of depth predictions by 9-64\% across a range of unseen, real-world validation datasets.
Importantly, we maintain consistency of all performance metrics across all experiments and scene depths where current vision-only approaches fail.
We further address the present deficit of training data in mobile robotics environments by introducing a novel methodology for synthesizing rendered, realistic learning datasets based on photogrammetric data that simulate the radar sensor observations for training.
Our code, datasets, and pre-trained networks are made available at \url{https://github.com/ethz-asl/radarmeetsvision}.
\end{abstract}

\section{Introduction}
Understanding the geometric structure of the environment is fundamental for autonomous robotics applications.
For instance, navigation in unknown environments requires an accurate, metric 3D representation of the scene \cite{Pantic2023ObstacleAvoidance}. A wealth of existing sensor modalities, such as \ac{LiDAR}, \ac{TOF}, and stereo cameras, are commonly used.
Typical dense 3D \ac{LiDAR} sensors are relatively expensive and large, \ac{TOF} range is often limited to a few meters, and stereo cameras need tight calibrations and correspondence matching \cite{8621614, 9127855}.
Recently, affordable single-chip \ac{FMCW} radars have been utilized for depth measurement in the automotive and aerial robotics domains \cite{9318758, girod2024brio}.
However, the output of such mmWave radar chips is typically extremely sparse and is subject to significant measurement noise. While this can be somewhat alleviated by accumulation and spatial alignment of radar data over time \cite{girod2024brio}, the resolution, density, and quality of the data are still orders of magnitude below typical \acp{LiDAR} or stereo setups.

\begin{figure}[t]
    \centering
    \includegraphics[width=0.45\textwidth]{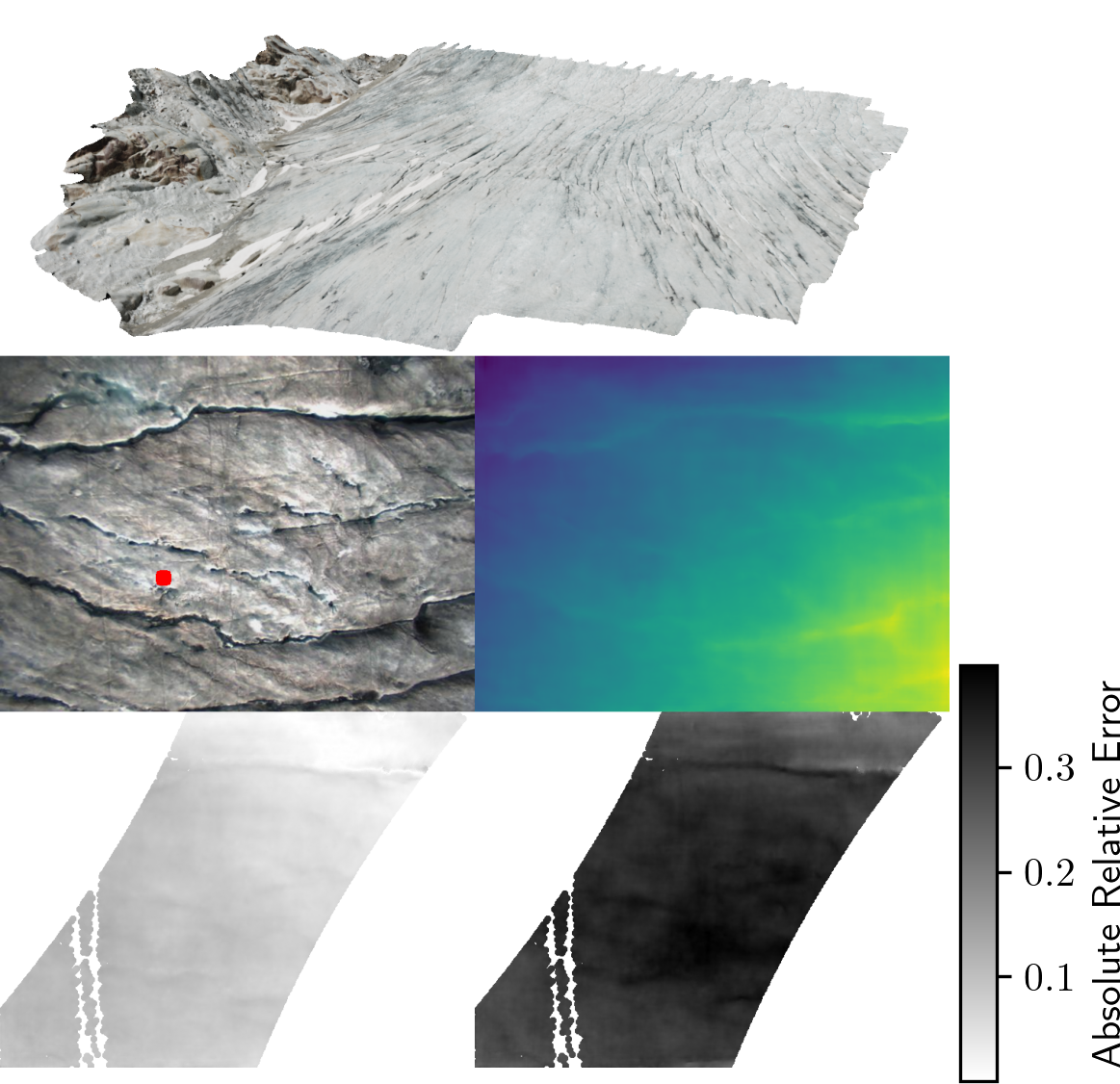}
    \caption{Top row: 3D rendering of the Rhône glacier in Switzerland, one of the validation testing sites. Middle row: RGB input image into the network, combined with the sparse radar observation in red (left) and metric depth prediction of our approach (right). Bottom row: Absolute relative error compared to LiDAR ground truth of our approach (left) and the work of \cite{depthanythingv2}.}
    \label{fig:introduction}
\end{figure}

With recent advances in metric \ac{MDE}, even the simple and cheap monocular camera can be used as a depth sensor. These seminal works have shown surprisingly strong generalization and robustness across various contexts \cite{bhat2023zoedepthzeroshottransfercombining, depthanything, depthanythingv2}, especially for applications where large datasets exist, such as automotive and indoor settings and scenes/imagery found in social media. However, many mobile robots primarily operate in outdoor or industrial environments with a lack of scale features and an abundance of self-similar, ambiguous, or low texture conditions that are known to be challenging for vision-based approaches. Further, the amount of training datasets available for such settings is not comparable to the wealth of data from the general internet and social media context. \Cref{fig:related_work_error} summarizes the performance of recent \ac{MDE} works in different environments, showing a clear lack of robustness for outdoor applications.

This work contributes a method for leveraging pre-existing state-of-the-art \ac{MDE} models in a robust fashion, addressing the performance deficits and lack of training data particular to outdoor robotics contexts.
Our approach combines sparse, metric radar depth measurements with dense, high-resolution \ac{MDE} models.
Doing so, we observe a notable improvement in robustness and generalization capabilities, which is crucial for deploying such a system on a mobile robot in complex, real-world environments.
Our approach directly encodes radar observations into the input space, ensuring compatibility with many existing \ac{MDE} frameworks.
In this paper, we use the \ac{SOTA} model ``DepthAnythingV2"\cite{depthanythingv2}, published in June 2024.
Our contributions entail 
\begin{itemize}
    \item a novel architecture that fuses radar into the 
    \ac{MDE} model using a custom loss method tailored to sparse radar measurements,
    \item a new radar and RGB vision dataset needed to fine-tune and validate the proposed architecture,
    \item and, due to the difficulty in obtaining large-scale datasets, a method for data augmentation based on 3D rendering of photogrammetry data.
\end{itemize}

Additionally, we demonstrate the real-world applicability of our method through extensive validation, showcasing significant performance improvements and making our datasets and code publicly available to facilitate further research and practical applications.

\section{Related Work}
\begin{figure}[t]
    \begin{tikzpicture}
        \begin{axis}[
            ylabel={{Absolute Relative Error}},
            xtick={0.3, 0.9500000000000002, 1.3500000000000003},
            x tick style={color=gray},
            xticklabels={{Automotive, Indoor, Outdoor}},
            extra x ticks={0.5499999999999999, 1.2500000000000002},
            extra x tick labels={},
            extra tick style={grid=major, color=black},
            ymajorgrids=true,
            y grid style=dashed,
            x grid style={thick}
        ]
        \addplot[only marks] coordinates { (0.0,0.045) };
        \node[right] at (axis cs:0.0,0.0049999999999999906 ) {\cite{depthanythingv2}};
        \addplot[only marks] coordinates { (0.1,0.057) };
        \node[left] at (axis cs:0.1,0.092 ) {\cite{bhat2023zoedepthzeroshottransfercombining}};
        \addplot[only marks] coordinates { (0.2,0.085) };
        \node[right] at (axis cs:0.2,0.05000000000000001 ) {\cite{depthanything}};
        \addplot[only marks] coordinates { (0.30000000000000004,0.086) };
        \node[left] at (axis cs:0.30000000000000004,0.121 ) {\cite{li2024radarcamdepthradarcamerafusiondepth}};
        \addplot[only marks] coordinates { (0.4,0.105) };
        \node[right] at (axis cs:0.4,0.07 ) {\cite{bhat2023zoedepthzeroshottransfercombining}};
        \addplot[only marks] coordinates { (0.6,0.056) };
        \node[right] at (axis cs:0.6,0.015999999999999993 ) {\cite{depthanythingv2}};
        \addplot[only marks] coordinates { (0.7,0.077) };
        \node[left] at (axis cs:0.7,0.11199999999999999 ) {\cite{bhat2023zoedepthzeroshottransfercombining}};
        \addplot[only marks] coordinates { (0.8,0.15) };
        \node[right] at (axis cs:0.8,0.11499999999999999 ) {\cite{depthanything}};
        \addplot[only marks] coordinates { (0.9,0.186) };
        \node[left] at (axis cs:0.9,0.221 ) {\cite{bhat2023zoedepthzeroshottransfercombining}};
        \addplot[only marks] coordinates { (1.0,0.363) };
        \node[right] at (axis cs:1.0,0.328 ) {\cite{depthanything}};
        \addplot[only marks] coordinates { (1.1,0.5) };
        \node[left] at (axis cs:1.1,0.535 ) {\cite{depthanything}};
        \addplot[only marks] coordinates { (1.3000000000000003,0.794) };
        \node[right] at (axis cs:1.3000000000000003,0.754 ) {\cite{depthanything}};

        \end{axis}
    \end{tikzpicture}
    \caption{Absolute Relative Error in the works \cite{depthanythingv2, bhat2023zoedepthzeroshottransfercombining, depthanything, li2024radarcamdepthradarcamerafusiondepth} divided into the categories `Automotive', `Indoor' and `Outdoor'. `Automotive' clearly outperforms the other categories, especially the `Outdoor' category.}
    \label{fig:related_work_error}
\end{figure}
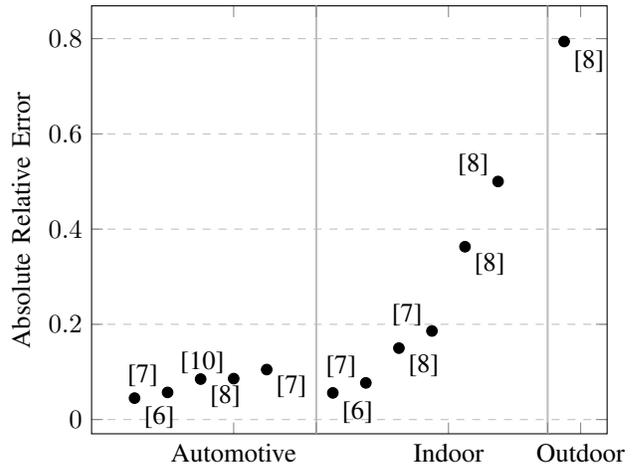

The field of pure monocular \ac{MDE} is evolving extremely fast - in the following, we present an overview of some of the most promising approaches as of September 2024. Due to the available training data, many of these works focus on automotive or indoor-like scenes. We give an overview of the absolute-relative error performance of the different works in various environments in \cref{fig:related_work_error}. 

A common backbone used in many (also non-metric) \ac{MDE} works is the MiDaS depth estimation framework~\cite{ranftl2020robustmonoculardepthestimation}. Combined with a metric binning module,~\cite{bhat2023zoedepthzeroshottransfercombining} achieves good performance for metric depth estimation for indoor and automotive environments.
While previous works used labeled data for training, the Depth Anything model~\cite{depthanything} enabled training on datasets with millions of unlabeled images and improved the encoder architecture, significantly increasing its zero-shot and detail performance. Especially in the latest follow-up work, \cite{depthanythingv2}, another improvement in metric depth estimation could be observed by adding the ability to train on synthetic data.
There are some limitations of solely relying on synthetic images, such as the domain gap between synthetic and real images, as well as limited scene coverage. To overcome these limitations, the authors apply a student-teacher approach, labeling real images using the most capable model to increase the dataset size. 

Besides pure vision-based \ac{MDE}, the idea of combining \ac{MDE} with radar is also being explored in the automotive sector. However, many works use multiple or high-end radars, often unsuitable for mobile robotics.
In~\cite{li2024radarcamdepthradarcamerafusiondepth}, a sensor setup consisting of up to five automotive-grade radars \cite{nuscenes} or a single high-end imaging radar that provides comparatively dense depth measurements as input to the depth estimation framework. The authors achieve a performance increase compared to previous works \cite{singh2023depth, DORN_radar, Gasperini_2021} by obtaining quasi-dense depth observations as an intermediate stage.

The approach of combining a monocular camera with a single, cheap, and lightweight \ac{FMCW} radar seems comparatively underdeveloped, likely owing to the highly sparse radar data, which is not compatible with the often used depth in-painting philosophy.

\section{Method}
Our method addresses this gap in the literature by customizing the original architecture DepthAnythingV2~\cite{depthanythingv2} for radar observations. We fine-tune it on semi-synthetic datasets that we obtain via image-based photogrammetry and validate our approach on data from a real camera-radar sensor system.

\subsection{Architecture}
\begin{figure*}[t]
    \centering
    \includegraphics[width=\textwidth]{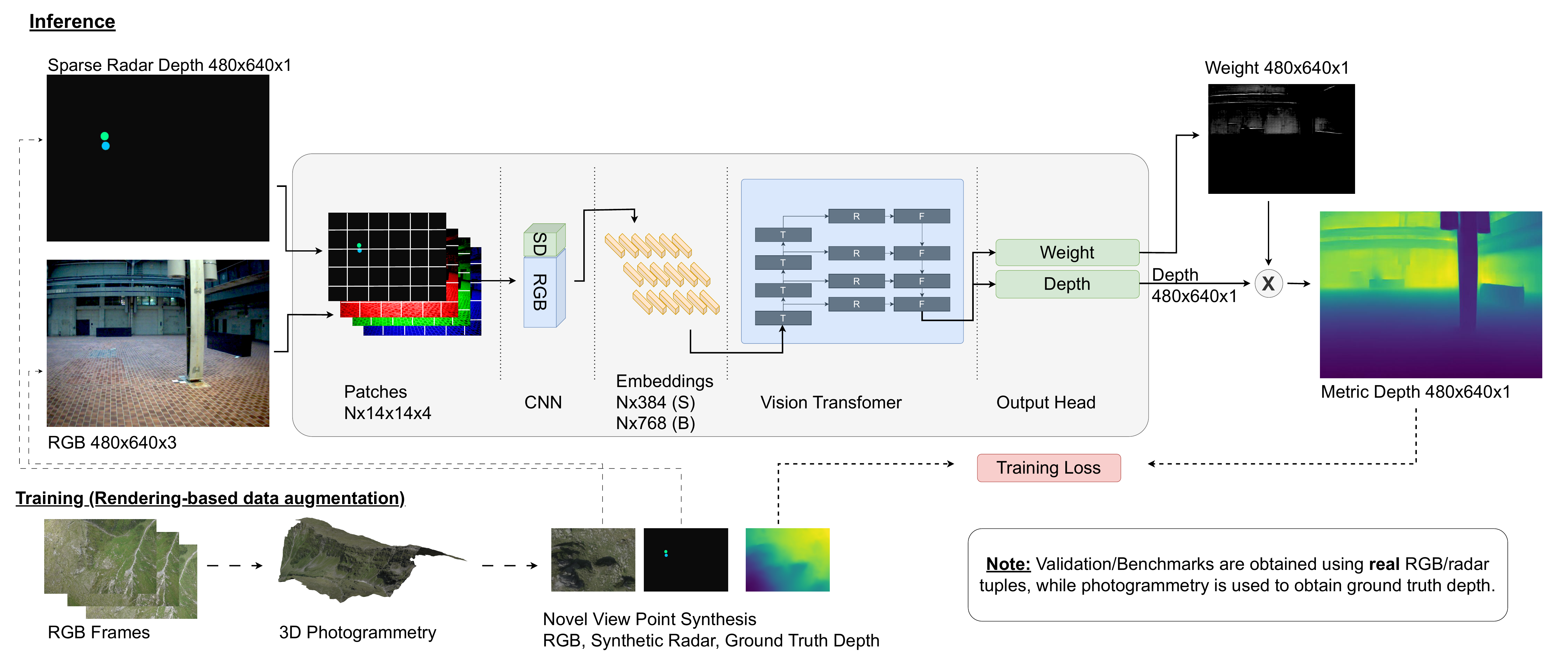}
    \caption{Overview of the inference and training architecture of our approach. We extend the input space of the architecture to $640\times 480 \times 4$; the additional channel encodes the sparse radar depth (\textit{SD}). We extend the network that creates the embeddings for the vision transformer to support the additional channel. The output head also extends to an additional channel, and we obtain the metric depth prediction described in \cref{eq:depth_weight}. All green components are trained at a high learning rate, whereas blue components are only fine-tuned.}
    \label{fig:blockdiagram}
\end{figure*}
The original DepthAnythingV2 architecture is designed and trained on RGB images only. In our approach, we extend the input space to facilitate the reuse of the weights present in the original model as much as possible and fine-tune them instead of re-training from scratch.
We project and render the radar data into a fourth channel besides the RGB data, as detailed in \cref{sec:radar_projection}. The resulting four-channel (R,G,B,radar) images are fed into a CNN that extracts the feature embeddings consumed by the vision transformer. We extend the previously present three-channel CNN by another input channel and augment the pre-trained weights with new untrained, random weights whenever needed. Otherwise, we retain the vision transformer architecture from \cite{depthanythingv2}. Accordingly, we fine-tune pre-existing weights in the CNN and vision transformer at a lower learning rate, whereas all added parameters are trained with a high learning rate.

We add a second output channel that represents the sigmoid-normalized pixel-wise weight $\mathbf{w}$ of the prediction of the network and the average of all $N$ radar observations $d_{i, radar}$.
The final depth prediction is then obtained via a combination of the two output channels:
\begin{equation}
    \hat{\mathbf{d}} = \hat{\mathbf{d}}_0 \cdot \mathbf{w} + (1.0 - \mathbf{w}) \cdot \frac{1}{N}\sum_{i=0}^{N-1}{d_{i, radar}}
\label{eq:depth_weight}
\end{equation}
where $\hat{\mathbf{d}}_0$ corresponds to the depth prediction of the depth output head and $\hat{\mathbf{d}}$ to the final metric depth prediction.

Intuitively, for each pixel, a higher weight trusts the output of the depth head more; a lower weight falls back to the averaged depth value from all radar observations in the image frame.
This mechanism ensures that the radar observations are incorporated when propagating the loss through the network. The idea is that the model learns to increase the weight when it performs better than purely utilizing radar observations.

The augmented depth output described in \cref{eq:depth_weight} is then used in a scale-invariant log loss function as in \cite{bhat2023zoedepthzeroshottransfercombining}.

\subsection{Training Datasets}
\begin{figure}[t]
    \centering
    \includegraphics[width=0.475\textwidth]{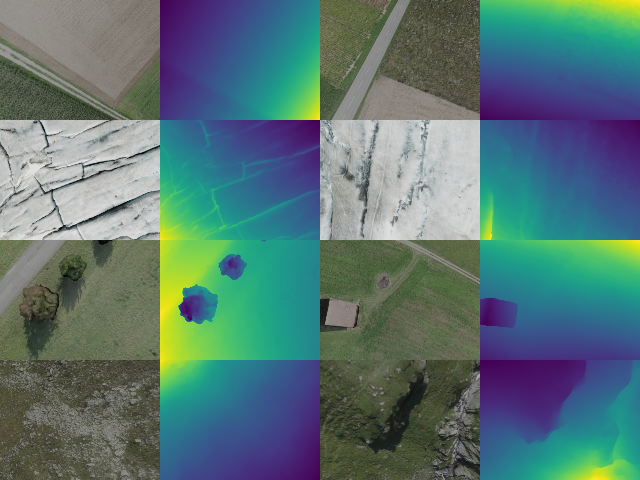}
    \caption{An overview of the four generated training datasets: From top to bottom row, we show samples of the road area, the Rhône glacier, the rural farming area, and the mountainous area.}
    \label{fig:dataset_overview}
\end{figure}

Similar to the original DepthAnythingV2, we train our network on synthetic data only. The amount of data needed for even just fine-tuning the vision transformer module surpasses the available body of calibrated image-radar data.
We use pure image data from multiple aerial photogrammetry datasets and obtain a 3D mesh using a commercially available, high-accuracy photogrammetry software (\textit{Pix4D}) that solves the \ac{SfM} problem in the area of interest. The datasets depict a typical road area, a high-altitude glacier, a rural farming area, and a mountainous area. Especially the datasets in nature are challenging for \ac{MDE}, as they contain a vast range of scale, absent or self-similar texture, and essentially do not contain artificial objects that allow for scale observation.

The training dataset, consisting of RGB images, sparse radar data rendered into the fourth channel, and the ground truth scene depth, is then created by generating random camera positions and attitudes on the 3D mesh. We randomly sample the camera's horizontal position to be within the extent of the mesh area, whereas the vertical position is uniformly sampled in $[1,51]$\,\unit{\metre} above the mesh surface.
Initially, the camera's z-axis points downwards, aligned with gravity on the mesh (see Fig. \ref{fig:handheld}). The camera's orientation is then randomly sampled between $\pm22.5\,^\circ$ for the camera's x- and y-axis, and the z-axis is sampled over the $\pm180\,^\circ$ range. We discard an orientation if the mesh is not entirely in the camera's view, as this leads to infinite depth.

We match the camera intrinsics to the intrinsics of the camera used in the experiments.
We then use \textit{Blender} to render these views in RGB and depth. The resulting synthetic images are of high quality, as is visualized in \cref{fig:dataset_overview}. We obtain the ground truth depth maps through ray tracing in the same step.

We use the depth ground truth as a supervision signal during training and dynamically generate synthetic radar observations from the ground truth depth at each training step. To do so, we detect 50 corner features in the rendered RGB image (using \cite{323794}) and randomly sample between 1 and 5 of these points as radar depth returns.
This approach dramatically increases the variability and dataset size of the radar information.
We choose 5 points as this is the maximum number of points observed in our experiments, after filtering by \ac{SNR} as described in \cref{sec:radar_projection}.

The intuition behind using corner features comes from the principle of radar cross-section, where corners often reflect radar signals more strongly. While more sophisticated models for radar simulation exist, this simple model was sufficient to train the network and allowed us to use any image-based dataset. The obtained synthetic radar observations are then rendered into a one-channel image identically to the real radar data, as explained in \cref{sec:radar_projection}.
We create four training datasets with RGB and depth, containing 10'000 samples each.

In addition, we use 10'000 samples of the HyperSim \cite{roberts:2021} dataset to train on indoor scenes as well, which we augment with synthetic radar observations in the same manner. 

\subsection{Validation Datasets}
\label{subsection:validation_datasets}
\begin{figure}[t]
    \centering
    \includegraphics[width=0.25\textwidth]{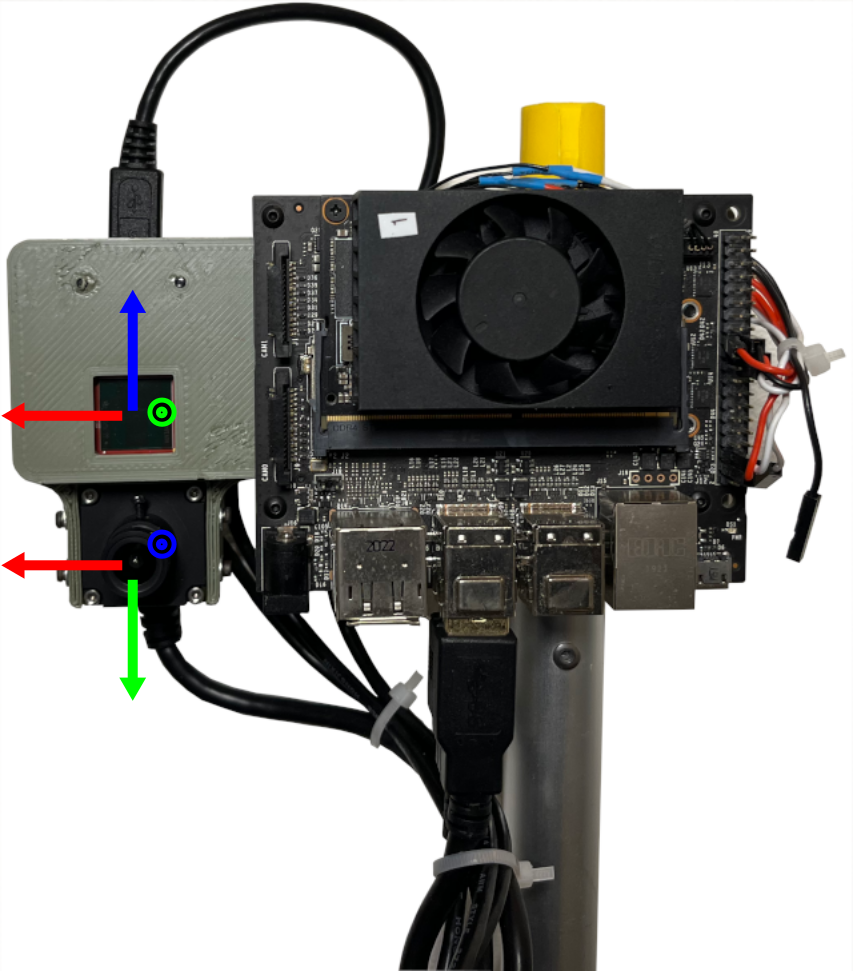}
    \caption{Handheld sensor rig, using a TI mmWave AWR1843AOPEVM radar, a FLIR FFY-U3-16S2C-S using a 3.6\,mm lense. Red arrows represent the x-axis, green arrows represent the y-axis, and blue arrows represent the z-axis. The data is recorded and processed by a Nvidia Jetson Xavier NX.}
    \label{fig:handheld}
\end{figure}

We collect a total of three diverse validation datasets. The \textit{Industrial Hall} and \textit{Agricultural Field} were obtained using a custom-built sensor rig that collects RGB images and radar observations, as shown in \cref{fig:handheld}. 
The \textit{Rhône Glacier} was collected on a \ac{MAV} flying over a high-altitude alpine glacier.

We mount the hand-held sensor rig on an up to 3m extendable pole to record diverse perspectives. The rig consists of a FLIR FFY-U3-16S2C-S global shutter camera with a maximum \SI{1.6}{MP} resolution, and a TI mmWave AWR1843AOPEVM radar. Both sensors, in an unsynchronized fashion, record at $20$\,\unit{\hertz}, and interface over USB. We reduce the output resolution using binning and a region of interest to $960\times1280$\,pixels. This resolution results from a trade-off between sufficient image resolution for photogrammetry and exactly being four times larger than the final input resolution to the network.
In addition to the two primary sensors, an Analog Devices ADIS16448 \ac{IMU} is mounted, recording at $200$\,\unit{\hertz}.
On both handheld datasets, in addition to the data collection rig, a DJI Mini 2 \ac{MAV} equipped with a \SI{12}{MP} RGB camera is used to collect images of the same area. The primary purpose for using the \ac{MAV} in addition is to increase the coverage of the area and provide a higher viewpoint. Especially in the case of the \textit{Agricultural Field} dataset, the nadir view improves the quality of the photogrammetry by providing \ac{GNSS} geolocation information included in the images.
In the case of the \textit{Industrial Hall} dataset, we provide absolute scale manually using a known calibration object. To ensure that both image sources can be combined, we place four manual tie points throughout each scene.

In order to make the computational load of the \ac{SfM} optimization problem more feasible, we subsample the images obtained by the handheld sensor rig by a factor of 3.
Using all images, manual tie points, and scale information, \textit{Pix4D} solves the \ac{SfM} problem for all intrinsics, extrinsics, and a sparse point cloud containing all 3D image features used.

We collect the validation dataset on the \textit{Rhône Glacier} on a heavily modified DJI M600 platform.
This \ac{MAV} uses the same sensors and setup as described above, and in addition, it collects \ac{LiDAR} observations using an Ouster OS1 sensor. \cref{tab:sparsity_summary} describes the number of sparse radar depths, the percentage of the image where the sparse ground truth is available, and the total number of frames used for validation.
\renewcommand{\thetable}{\arabic{table}}
\captionsetup[table]{labelformat=simple, labelsep=colon, name=Tab.}
\begin{table}[h]
\centering
\sisetup{detect-all}
\NewDocumentCommand{\B}{}{\fontseries{b}\selectfont}
\begin{tabular}{@{}l S[table-format=1.2] S[table-format=2.2] S[table-format=3]@{}}
\toprule
Dataset & {\# Sparse Depth} & {\% Depth GT Coverage} & {\# Frames} \\ 
\midrule
Industrial Hall & 2.96 & 37.80 & 365 \\ 
Agricultural Field & 1.07 & 51.90 & 86 \\ 
Rhône Glacier & 2.37 & 3.32 & 302 \\ 
\bottomrule
\end{tabular}
\caption{The average number of sparse radar depth points per frame, average ground truth coverage per frame, and number of samples per dataset are shown.}
\label{tab:sparsity_summary}
\end{table}

Only the real radar observations obtained during the data collection are used for validation, and no radar measurements are synthesized. As a validation ground truth, we use bundle-adjusted \ac{SfM} or LiDAR point clouds backprojected into the image frame using the camera's intrinsics.

\subsection{Radar Image Projection}
\label{sec:radar_projection}
The radar outputs a sparse 3D point cloud (usually about 1-5 points) in Cartesian coordinates, including a \ac{SNR} value per point. In order to feed this data into the network, we project the sparse depth points onto a single channel image. The radar sensor internally utilizes the \ac{CFAR} algorithm that operates on the data coming from the radar front-end. The \ac{CFAR} algorithm performs a sort of non-maximum suppression, where it only keeps strong radar returns.
We filter the received radar observations by their \ac{SNR}, with a cutoff of $15\,dB$. This yielded a satisfying performance on all evaluated datasets.
We then accumulate the radar data over three frames ($\approx\SI{150}{\milli\second}$) and project the accumulated points onto the camera image plane using the general pinhole camera equation:
\begin{equation}
_{c}\mathbf{p}_{r} = K \left[R_{c, r} | t_{c,r}\right] _{r}\mathbf{P}_{r}
\end{equation}
where $\mathbf{P}_{r}$ is the homogeneous radar point cloud in the radar coordinate system, $R_{c, r}$ and $t_{c,r}$ are the rotation and translation from radar to camera frame respectively, $K$ the camera intrinsics and $_{c}\mathbf{p}_{r}$ the homogeneous projected points in image coordinates. We discard all points that lie outside of the image coordinates.
All projected radar points get padded with a 5-pixel radius circular shape, as shown in \cref{fig:blockdiagram}. We set the pixel value directly to the depth value of the corresponding radar observation and encode any pixels without radar observations in this channel as zeros.

This mapping requires the relative transformation matrix between the camera and the radar. We obtain the full radar to camera calibration $R_{c, r}, t_{c,r}$ via calibrating both sensors w.r.t to the \ac{IMU} as a shared reference. We calibrate the camera intrinsics and extrinsics using \cite{kalibr}, while the radar-\ac{IMU} transformation is obtained using CAD. The geometric center of the radar antenna is the reference point.

\section{Experimental Design}
We evaluate our approach against the \ac{SOTA} baseline \cite{depthanythingv2} fine-tuned for metric estimation using our training datasets. We also provide another comparison to a ``naive" approach of scaling the relative output of the \ac{SOTA} baseline with the radar depth. The naive approach shows how much distortion is present in the output depth maps of modern \ac{MDE} models, as a simple re-scaling will not perform well if the depth prediction distortion is non-linear.
Additionally, we train and evaluate all approaches using two different network sizes, small (\textit{S}) and base (\textit{B}) (same nomenclature as \cite{depthanythingv2}). The main difference lies in the size of the embeddings: \textit{S} uses 384-wide embeddings, whereas \textit{B} uses 768. The \textit{B} model also doubles the number of attention heads to 12. \cref{tab:inference_summary} shows the total number of resulting parameters.

As the radar sensor often only returns a single observation (cf. \cref{tab:sparsity_summary}), we use a single scalar value $\hat{s}_d$ to scale the output depth map $\mathbf{d}$ of the naive approach.
The scalar $\hat{s}_d$ and the scaled metric depth $\hat{\mathbf{d}}$ are computed as the mean scaling factor
\begin{equation}
\begin{split}
    \hat{s}_d &= \frac{1}{N}\sum_{i=0}^{N-1}{\frac{d_{i, radar}}{\hat{d}_{i, rel}}}\\
    \hat{\mathbf{d}} &= \hat{s}_d \cdot \hat{\mathbf{d}}_{rel}
\end{split}
\end{equation}
where $N$ is the number of radar observations and $\hat{d}_{i,rel}$ the relative depth prediction at the image coordinates of the radar observation $d_{i,radar}$.

We train all networks in the same manner: The training terminates after 25 epochs with 50'000 training steps each.
All networks use pre-trained weights provided by the work \cite{depthanythingv2}, fine-tuned to the outdoor task on Virtual KITTI 2 \cite{cabon2020vkitti2}. We chose training batch sizes according to the available hardware (Nvidia RTX4090), resulting in batch sizes of 8 for the S-sized networks and 4 for the B-sized networks.
We apply a polynomial decay learning rate scheduler with a power of $0.9$ together with the Adam optimizer, starting from a learning rate of $5\cdot10^{-6}$ for pre-trained weights and ten times larger for high-learning rate weights. The learning rate monotonically decreases to zero at the end of all training steps, corresponding to the implementation in \cite{depthanythingv2}.
After training, we choose the weights from the epoch with the lowest validation \textit{AbsRel} for our final validation.
In all experiments, the performance stagnates towards the end of the 25 epochs.
\renewcommand{\thetable}{\arabic{table}}
\captionsetup[table]{labelformat=simple, labelsep=colon, name=Tab.}
\begin{table}[h]
\centering
\sisetup{detect-all}
\NewDocumentCommand{\B}{}{\fontseries{b}\selectfont}
\begin{tabular}{@{}lS[table-format=2.2]S[table-format=3.1]@{}}
\toprule
Models & {\# Parameters (M)} & {Avg. Inference Time (ms)} \\ 
\midrule
Metric Depth \cite{depthanythingv2}-B & 97.47 & 114.1 \\ 
Metric Depth \cite{depthanythingv2}-S & 24.79 & 43.5 \\ 
Ours-B & 97.62 & 112.5 \\ 
Ours-S & 24.86 & 42.8 \\ 
\bottomrule
\end{tabular}
\caption{Comparison of the number of model parameters and the inference time using our approach and Depth Anything V2. The suffix -S and -B indicates the pre-trained network size, small and base, respectively \cite{depthanythingv2}.}
\label{tab:inference_summary}
\end{table}

\section{Results}
In the following section, we will present the results and discuss the most important findings and their interpretation.
\renewcommand{\thetable}{\arabic{table}}
\captionsetup[table]{labelformat=simple, labelsep=colon, name=Tab.}
\begin{table*}[!ht]
\centering
\sisetup{detect-all}
\NewDocumentCommand{\B}{}{\fontseries{b}\selectfont}
\def\Decimal{.000}
\def\Uline#1{\Ulinehelp#1 }
\def\Ulinehelp#1.#2 {%
  #1.#2\setbox0=\hbox{#1\Decimal}\hspace{-\wd0}{\if\relax#2\relax%
    \uline{\phantom{#1.0}}\else\uline{\phantom{#1.#2}}\fi}%
}
\begin{tabular}{
@{}
l
S[table-format=2.3]
S[table-format=2.3]
S[table-format=2.3]
S[table-format=2.3]
S[table-format=2.3]
S[table-format=2.3]
S[table-format=2.3]
S[table-format=2.3]
S[table-format=2.3]
@{}
}
\toprule
 & \multicolumn{3}{c}{Industrial Hall} & \multicolumn{3}{c}{Agricultural Field} & \multicolumn{3}{c}{Rhône Glacier} \\
\cmidrule(lr){2-4} \cmidrule(lr){5-7} \cmidrule(lr){8-10}
Models & {AbsRel $(\downarrow)$} & {$\delta_1 (\uparrow)$} & {RMSE $(\downarrow)$} & {AbsRel $(\downarrow)$} & {$\delta_1 (\uparrow)$} & {RMSE $(\downarrow)$} & {AbsRel $(\downarrow)$} & {$\delta_1 (\uparrow)$} & {RMSE $(\downarrow)$} \\
\midrule
Metric Depth \cite{depthanythingv2}-S & 0.206 & 0.485 & 2.231 & 3.750 & 0.012 & 21.800 & 3.827 & 0.000 & 19.619 \\
Metric Depth \cite{depthanythingv2}-B & \Uline 0.194 & 0.587 & \Uline 2.197 & 1.632 & 0.067 & 9.639 & 2.669 & 0.000 & 13.666 \\
Naive-S & 1.959 & 0.211 & 18.449 & 0.872 & 0.136 & 8.941 & 0.292 & 0.397 & 2.222 \\
Naive-B & 2.705 & 0.155 & 27.982 & 0.952 & \Uline 0.139 & 9.473 & \Uline 0.247 & 0.467 & 1.878 \\
Ours-S & 0.235 & \Uline 0.612 & 2.467 & \Uline 0.463 & 0.136 & \Uline 6.608 & 0.272 & \Uline 0.532 & \Uline 1.565 \\
Ours-B & \B 0.170 & \B 0.709 & \B 2.120 & \B 0.313 & \B 0.331 & \B 4.916 & \B 0.223 & \B 0.686 & \B 1.436 \\
\bottomrule
\end{tabular}
\caption{Comparison of metric Depth Anything V2 \cite{depthanythingv2} and the naive approach with our system. The suffix -S and -B indicates the pre-trained network size, which is small and base, respectively. The best values are in bold, and the second-best values are underlined. 
\textit{AbsRel} is the metric absolute relative error, $\delta_1$  the thresholded accuracy (i.e., $\max{(\mathbf{d}/\hat{\mathbf{d}}, \hat{\mathbf{d}}/\mathbf{d})}<1.25$) and RMSE the root mean square error in meters.
}
\label{tab:results}
\end{table*}
\begin{figure*}[!ht]
    \centering
    \includegraphics[width=0.95\textwidth]{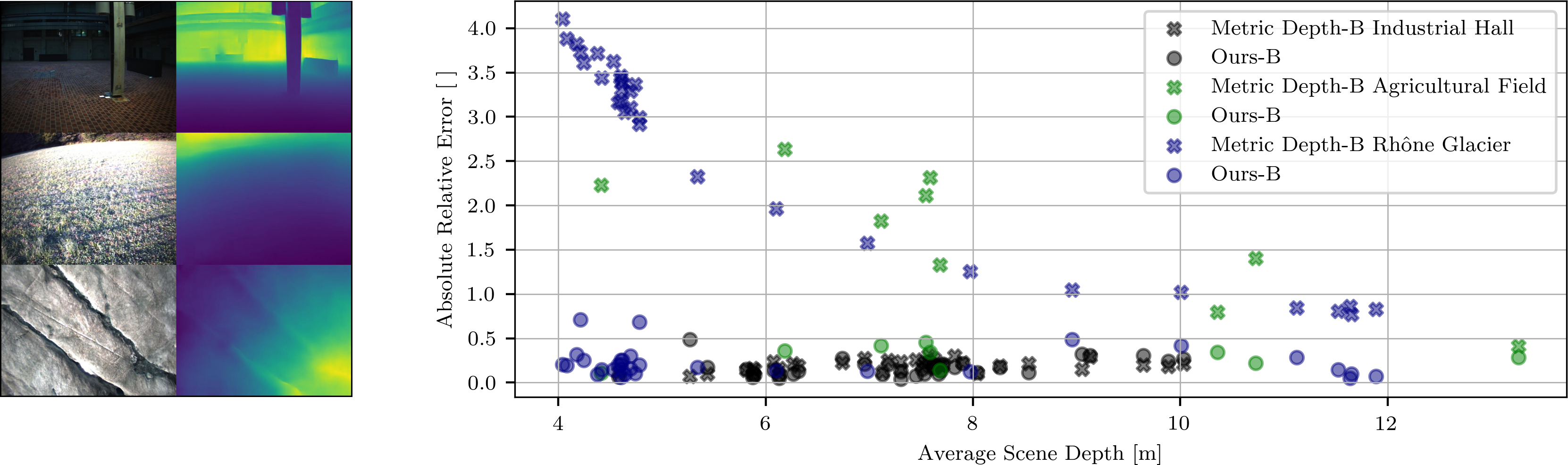}
    \caption{Left: RGB input and corresponding depth prediction using \textit{Ours-B} for each of our validation datasets \textit{Industrial Hall}, \textit{Agricultural Field}, \textit{Rhône Glacier}. Right: Absolute relative error over the average scene depth for each dataset and network. Each plot point represents one dataset frame; we subsample the dataset by a factor of 10 for visualization purposes. For simplicity, only \cite{depthanythingv2} and Ours, using the base-sized networks, are shown.}
    \label{fig:results_overview}
\end{figure*}
\Cref{tab:results} shows an overview of the quantitative results of all methods on three different datasets.
\textit{Ours-B} outperforms all other models in all three experiments, with the base network size generally outperforming the smaller network.
There are single metrics where the small-size network performs very closely, or even better than the base-size network, i.e., on the experiment \textit{Agricultural Field} with the naive approach on \textit{AbsRel} and RMSE.  
The pure metric \ac{MDE} approach performs comparatively well in the \textit{Industrial Hall} dataset, likely due to its loose similarity to non-robotics-oriented indoor datasets used in training. However, on more difficult out-of-distribution datasets such as the \textit{Agricultural Field} or \textit{Rhône Glacier}, the missing generalization of the pure metric \ac{MDE} approach becomes apparent. The performance advantage of our approach versus the naive approach also shows that the undistortion of depth estimations benefits from tight fusion and that our approach seems to increase the model accuracy while considering radar data.

Our approach's key performance improvement over the baselines is the consistent accuracy across all experiments, regardless of the environment. \Cref{fig:results_overview} relates the absolute relative error with the ground truth distance of the baseline and our method, showcasing the depth-dependent error distribution over multiple datasets.

Examining the agricultural and glacier datasets, we observed a tendency for the absolute relative error to increase towards lower scene depths. Intuitively, this shows that the baseline metric network systematically under- or over-estimates specific validation dataset frames.
Our approach's absolute mean error stays relatively constant across the whole depth range, confirming that our approach successfully incorporates scale information and that the error is not strongly tied to the scale of the scene. 

Overall, the datasets used in this evaluation are challenging and representative of typical applications of mobile robots. All validation datasets contain, to some degree, unknown environments, challenging lighting conditions, self-similarity, and ambiguous scales, as is visualized qualitatively in \cref{fig:results_overview} on the left. One limitation of our approach is handling depth at infinity, such as horizons, which it shares with many \ac{MDE} approaches. However, the presented approach successfully provided robust and precise depth estimation even for complex scenes completely outside the training distribution, making it robust enough for mobile robotics applications. 
\section{Conclusion}
This work presents a novel approach for metric depth prediction in unknown environments. Our model makes the combination of a monocular camera and a low-cost mmWave radar a viable dense metric depth sensor modality for mobile robotics and outperforms the baselines in standard depth prediction performance metrics; we observe improvements of $9$-$64\,\%$ in absolute relative error. Most importantly, the approach performed consistently on a diverse array of datasets. The so-obtained robustness in depth perception is crucial for mobile robotics, where any defects in the estimated data may have considerable implications on the robot's safety.

Additionally, we contribute a method for generating large amounts of training data for sparse radar-based methods. Doing so, we drastically lower the need for manual data collection and simultaneously circumvent issues that may arise when training on noisy and sparse radar data, as reported in  \cite{li2024radarcamdepthradarcamerafusiondepth}.
In the future, we plan to use the presented approach to replace typical depth sensors in downstream tasks such as collision avoidance or mapping.

\printbibliography
\end{document}